\documentclass[journal]{IEEEtran}

\usepackage{cite}

\ifCLASSINFOpdf
  \usepackage[pdftex]{graphicx}
\else
  \usepackage[dvips]{graphicx}
\fi
\usepackage{amsmath}
\usepackage{algorithmic}
\usepackage[linesnumbered,ruled]{algorithm2e}

\usepackage{array}

\ifCLASSOPTIONcompsoc
  \usepackage[caption=false,font=normalsize,labelfont=sf,textfont=sf]{subfig}
\else
  \usepackage[caption=false,font=footnotesize]{subfig}
\fi
\usepackage{fixltx2e}

\usepackage{stfloats}
\newcolumntype{P}[1]{>{\centering\arraybackslash}p{#1}}

\usepackage{booktabs}

\usepackage{xcolor}

\usepackage{comment}
\usepackage{threeparttable}
\newcommand{\tabincell}[2]{\begin{tabular}{@{}#1@{}}#2\end{tabular}}

\begin{document}
%
\title{Explaining Deep Convolutional Neural Networks for Image Classification by Evolving Local Interpretable Model-agnostic Explanations}
%
%
%

\author{Bin Wang,
		Wenbin Pei,~\IEEEmembership{Member,~IEEE,}
		Bing Xue,~\IEEEmembership{Senior Member,~IEEE,}
        Mengjie Zhang,~\IEEEmembership{Fellow,~IEEE,}
		\thanks{Corresponding author: Wenbin Pei.}
		\thanks{© 2025 IEEE.  Personal use of this material is permitted.  Permission from IEEE must be obtained for all other uses, in any current or future media, including reprinting/republishing this material for advertising or promotional purposes, creating new collective works, for resale or redistribution to servers or lists, or reuse of any copyrighted component of this work in other works.}
}

%
%

\markboth{JOURNAL OF LATEX CLASS FILES, VOL. XX, NO. XX, March 2025}%
{Shell \MakeLowercase{\textit{et al.}}: Bare Demo of IEEEtran.cls for IEEE Journals}
%



\maketitle

\begin{abstract}

Deep convolutional neural networks (CNNs) have proven their effectiveness and are widely acknowledged as the dominant method for image classification. However, their lack of explainability remains a significant drawback, particularly in real-world applications where users need to understand the rationale behind predictions to determine their trustworthiness. Local Interpretable Model-agnostic Explanations (LIME) is a popular method for explaining deep CNN predictions, but it suffers from two major limitations: (1) a computationally expensive sampling process to generate perturbed images, and (2) the need to pre-define the number of interpretable features (superpixels), which often requires manual fine-tuning. To address these limitations, we propose a novel Genetic Algorithm (GA)-based method, called E-LIME, which automatically evolves local explanations for deep CNNs. The proposed method eliminates the need for the computationally expensive sampling process used in LIME and allows for the automatic selection of superpixels without pre-defining their number. Specifically, E-LIME introduces a flexible encoding strategy to represent superpixels as binary vectors and a new fitness function that evaluates the selected superpixels based on the probability of the deep CNN making a specific prediction. By optimising the fitness value, E-LIME selects the most informative superpixels while removing noisy features, resulting in more efficient and accurate local explanations. In the experiments, ResNet is used as an example model to be explained, and the ImageNet dataset is selected as the benchmark dataset. DenseNet and MobileNet are further explained to demonstrate the model-agnostic characteristics of the proposed method. The evolved local explanations on four randomly selected images from ImageNet show that the proposed method successfully captures meaningful interpretable features, improving the probabilities/confidences of the deep CNN models in making predictions. Moreover, the proposed method obtains local explanations within one minute, which is more than ten times faster than LIME. The proposed E-LIME method not only overcomes the limitations of LIME but also provides a more efficient and flexible approach to explaining deep CNN predictions, making it highly suitable for real-world applications where interpretability and computational efficiency are critical.
\end{abstract}


%

\section{Introduction}
\IEEEPARstart{D}{eep} convolutional neural networks (CNNs) have achieved state-of-the-art classification accuracies on image classification tasks \cite{wu2019automatic, nie20183, behera2021crowd, xiao2021pam, lan2023compact}, often surpassing human performance. Over the past decade, a variety of deep CNN architectures have been proposed, including VGGNet \cite{simonyan2014very}, ResNet \cite{he2016deep}, DenseNet \cite{huang2017densely}, and MobileNetV2 \cite{sandler2018mobilenetv2}. Despite their remarkable performance, these models are often considered black boxes due to their complexity, making it difficult for humans to understand the rationale behind their predictions. This lack of explainability is particularly problematic in critical applications such as medical diagnosis, autonomous driving, and law enforcement, where understanding the decision-making process is essential for trust and accountability.

To address this issue, researchers have focused on developing methods for interpretability and explainability in machine learning \cite{dhebar2020interpretable, zhang2021transductive, wang2021interpretability, zhang2021survey, kiani2022towards, heuillet2022collective}. A prominent area of research, known as \textit{Explainable Deep Learning} (XDL) \cite{ghosal2018explainable, townsend2019extracting, ahn2020explaining, samek2021explaining, dhebar2022toward}, aims to provide insights into the behaviour of black-box deep learning models. XDL methods can be broadly categorised into two branches: \textit{model translation} and \textit{local approximation}.

\begin{itemize}
    \item \textbf{Model Translation}: These methods aim to globally approximate the behaviour of deep CNNs across the entire dataset using simpler, interpretable models such as decision trees \cite{frosst2017distilling, zhang2019interpreting}. However, this approach often sacrifices classification accuracy, as deep CNNs are too complex to be fully captured by simpler models.
    \item \textbf{Local Approximation}: These methods focus on providing explanations for individual predictions. Local Interpretable Model-agnostic Explanations (LIME) \cite{ribeiro2016should} is the state-of-the-art local approximation method. LIME segments an image into superpixels \cite{achanta2012slic} and identifies a subset of superpixels that are most influential in the model's prediction. While effective, LIME has two major limitations: (1) it requires a computationally expensive sampling process to generate perturbed images, and (2) the number of superpixels must be pre-defined, which often requires manual fine-tuning.
\end{itemize}

Evolutionary Computation (EC) has shown its effectiveness in automatically designing deep CNNs from individual EC methods \cite{wang2018evolving} \cite{miikkulainen2019evolving}\cite{wang2020particle} \cite{dong2022cell} \cite{liu2020evolving} \cite{deng2023evolutionary} to hybrid EC methods \cite{wang2018hybrid} \cite{wang2019hybrid} \cite{chen2021cde} \cite{fang2023lonas}, and from single-objective EC methods \cite{wang2019particle} \cite{lawrence2021particle}\cite{wang2020surrogate} \cite{wei2021self} \cite{lin2022evolutionary} to multi-objective EC methods \cite{elsken2018efficient} \cite{wang2019evolving} \cite{liang2020multiobjective} \cite{wen2021two}.  However, no exploration has been made by using EC to learn explanations of deep CNNs. Inspired by LIME, a novel Genetic Algorithm (GA) based method, to overcome the above two limitations of LIME, is proposed as a local approximation method of XDL.

As one of the most significant evolutionary algorithms, GA has gained a great deal of interest because of its broad applications to many different tasks, such as neural architecture search and complex optimization problems. In GA, a population of candidate solutions can be automatically evolved without explicitly telling them specific steps. The goodness of every solution is evaluated by a fitness function, and then fitness values are used to select good solutions by a selection operator for generating potentially better solutions. This contributes to GA-based methods being independence of complex domain knowledge. It is noted that LIME still needs domain knowledge to determine which types of explanations to be used and trained, such as linear models or decision trees. The use of GAs for learning explanations can avoid this. 

To sum up the rationale for using GA in our method is based on the combinatorial nature of the problem and the need for efficient optimisation. The task of selecting the most relevant superpixels to explain a deep CNN's prediction can be framed as a combinatorial optimisation problem. GA is well-suited for solving such problems due to its ability to explore a large search space efficiently and find near-optimal solutions. Additionally, the flexibility of GA in selecting any number of superpixels and its ability to eliminate the need for the sampling process used in LIME make it an ideal choice for evolving local explanations. The fitness function in E-LIME is designed to evaluate the selected superpixels based on the output of any deep CNN model, ensuring that the explanations are tailored to the specific model being explained.
\subsection{Contributions}
The overall goal of this paper is to propose a novel GA-based method to evolve local interpretable model-agnostic explanations. The proposed method (named E-LIME) is targeted at overcoming the aforementioned limitations of LIME, i.e., the high computational cost and the fixed number of superpixels. To the best of our knowledge, E-LIME is the first EC-based method to evolve local interpretable model-agnostic explanation for deep CNNs in image classification. By achieving the overall goal, the specific contributions are made as follows: 
\begin{enumerate}

\item To develop a novel GA-based method that can efficiently evolve local explanations. Since LIME uses a linear model with Lasso regularisation \cite{efron2004least} to select superpixels, it requires high computational cost for sampling a large number of perturbed images. To resolve this issue, the proposed method will directly optimise the superpixel selection  without any sampling process. 
	
\item To design a new encoding strategy that can encode the interpretable features, i.e., superpixels, to a binary vector. The encoding strategy needs to accommodate all of the superpixels, so it is possible for the proposed method to select any number of superpixels. However, LIME needs to define the number of superpixels beforehand, which may require fine-tuning for different images. The proposed method will break LIME's limitation of pre-fixing the number of selected superpixels.

\item To propose a new fitness function that can effectively evaluate the selected superpixels. The proposed fitness function is expected to have the ability of evaluating the selected superpixels represented by the encoded vector based on any deep CNN model. By optimising the fitness value, the superpixels that improve the probability of deep CNNs for making the specific prediction are selected, while the noisy superpixels are removed. Therefore, the proposed method can be leveraged to explain the predictions of any deep CNN model. 

\item To perform further analysis to examine the explainability and the model-agnostic characteristic of the proposed method. These analyses confirm that the proposed method has the ability to explain the predictions of deep CNN models without knowing the details of the deep CNNs.

\end{enumerate}

\subsection{Organisation}
The paper is organised as follows. In Section \ref{SSS:elime_background}, the essential background is introduced, and the related works are reviewed. Section \ref{SSS:elime_method} introduces the proposed method. The experiment setup is introduced in Section \ref{SSS:elime_experiment}, and the results are shown and discussed in Section \ref{SSS:elime_result}. Section \ref{SSS:elime_conclusion} concludes the paper. 

\section{Background and Related Works} \label{SSS:elime_background}

In this section, 
a widely-used method of generating super-pixels is introduced, which is used by the proposed method to produce interpretable features. Next, a deep CNN model known as ResNet is described as an example to explain the black-box characteristic of deep CNNs. This is also the motivation of why it is necessary to explain the predictions of deep CNNs. Finally, a brief literature review of existing work related to explaining deep CNNs is given.  

\subsection{Simple Linear Iterative Clustering (SLIC) Superpixels}
In image analysis, a group of pixels is regarded as a superpixel, which provides a convenient and compact representation of images. The use of superpixels could be useful for tasks that are computationally expensive, and it significantly reduces the complexity of subsequent image processing tasks \cite{achanta2012slic}. For generating superpixels, many algorithms have been designed and proposed, such as graph-based and gradient-ascent-based algorithms \cite{moore2008superpixel, veksler2010superpixels, vincent1991watersheds}. Simple Linear Iterative Clustering (SLIC) is considered as one of the most popular and efficient superpixel generation algorithms \cite{ribeiro2016should}\cite{ribeiro2018anchors}. 

SLIC generates superpixels based on $k$-means clustering, i.e., pixels are grouped based on their colour proximity and spatial proximity. In more detail, a cluster centre is represented as a $5 \times 1$ matrix $C_i=[l_i\;a_i\;b_i\;x_i\;y_i]^T$, where $l_i$, $a_i$ and $b_i$ are from the pixel colour vector in the CIELAB colour space, and $x_i$ and $y_i$ indicate the pixel positions. After an initialisation of $k$ cluster centres $C_k$ by sampling pixels at regular grid intervals $S$ ($S=\sqrt{\frac{N}{k}}$, where $N$ is the number of pixels), the centres move to seed locations with the lowest gradient position in a $3\times 3$ neighbourhood. The purpose is to avoid sampling noisy pixels or pixels located on an edge. Afterwards, for each cluster centre $C_k$, SLIC calculates the distance $D$ between pixel $i$ and $C_k$, to assign pixel $i$ to its nearest cluster. After each iteration, SLIC updates cluster centres and a residual error $E$. The whole process ends when $E \leq threshold$, which is a pre-defined value.

When calculating Euclidean distances, the spatial proximity outweighs the colour proximity for the larger superpixels because the image size influences the spatial distance. This gives a higher priority to the spatial proximity than the colour proximity. Hence, a normalized distance measure that considers the superpixel size is defined \cite{achanta2012slic}:
\begin{equation}\label{eqn:split}
D= \sqrt {d_c^2+(\frac{d_s}{S})m^2}
\end{equation}
where $d_c=\sqrt{(l_j-l_i)^2+(a_j-a_i)^2+(b_j-b_i)^2}$, $d_s=\sqrt{(x_j-x_i)^2+(y_j-y_i)^2}$, $S$ is a grid interval, and $m$ is used to weigh the relative importance between the colour similarity and the spatial proximity ($m$ can be in the range of [1, 40] when using the CIELAB colour space \cite{achanta2012slic}).

\subsection{Genetic Algorithm (GA)}
The Genetic Algorithm (GA) is a population-based optimisation technique inspired by natural selection. GA evolves a population of candidate solutions over generations to find optimal or near-optimal solutions. The process involves the following steps:
\begin{itemize}
    \item \textbf{Initialisation:} A population of candidate solutions (individuals) is randomly generated, typically encoded as binary strings or real-valued vectors.
    \item \textbf{Selection:} Individuals are selected based on their fitness, which measures how well they solve the problem. Common selection methods include tournament and roulette wheel selection.
    \item \textbf{Crossover:} Selected individuals (parents) are combined to produce offspring, using operators like single-point or uniform crossover.
    \item \textbf{Mutation:} Offspring undergo random changes (e.g., bit-flip or Gaussian mutation) to introduce diversity.
    \item \textbf{Replacement:} The new population is formed by replacing some or all of the previous generation with the offspring. This process repeats until a termination criterion is met, such as reaching a maximum number of generations or achieving a satisfactory fitness level.
\end{itemize}

In E-LIME, GA optimises the selection of superpixels by evaluating each candidate solution based on the probability of the deep CNN making a specific prediction. Through evaluation, selection, crossover and mutation (offspring generation), GA efficiently identifies the most informative superpixels, producing interpretable and accurate local explanations.

\subsection{ResNet}
\begin{figure}[ht]
	\centering
	\includegraphics[width=\linewidth]{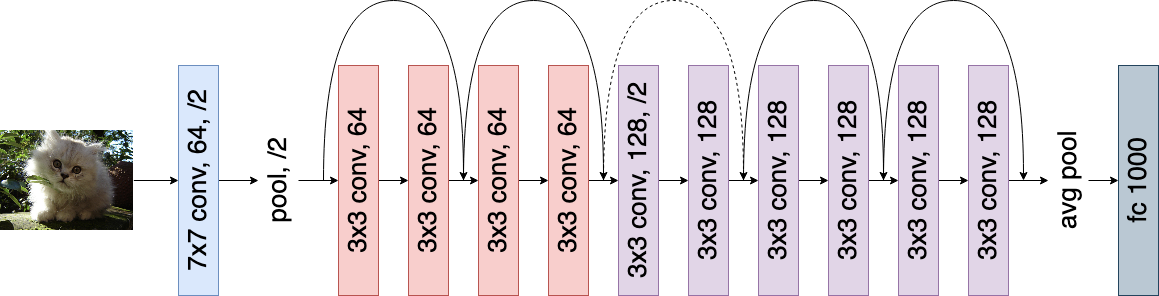}
	\caption{ResNet architecture \cite{he2016deep}.}
	\label{fig:cnn-blocks-resnet}
\end{figure}

ResNet \cite{he2016deep} is one of the state-of-the-art deep CNNs, which has achieved superior popularity in image classification, so ResNet is taken as the first of the three deep CNNs to be explained by the proposed method. Therefore, the  background of ResNet is briefed here. Fig. \ref{fig:cnn-blocks-resnet} displays a ResNet architecture with only 12 parameter layers (while in real-world applications, ResNet usually contains more than 100 parameter layers \cite{he2016deep}).  As shown in the figure, the input image is first passed to a convolutional layer with a filter size of $7\times7$ and a stride size of $2$, which produces 64 feature maps as the output feature maps. A pooling layer, which is a non-parameter layer, is used to downsample the output feature maps by $2$. Then, a group of 4 convolutional layers with a filter size of $3\times3$ are appended to further build 64 output feature maps.  After that, the output feature maps are then passed to another set of 6 convolutional layers with a filter size of $3\times3$ to construct 128 output feature maps. At the end, the output feature maps are downsampled by an average pooling layer, and then passed to a fully-connected layer to perform the final prediction of the input image's label. One notable characteristic of ResNet is the residual connection. The curves with arrows in Fig. \ref{fig:cnn-blocks-resnet} represent the residual connections, which pass the output feature maps of one layer to the layer after its next layer. The main purpose of the residual connections is to keep the learned feature maps available to deeper layers down the CNN architecture, which improves the performance by making the architecture easier to be trained \cite{he2016deep}.  

\begin{figure}[ht]
	\centering
	\begin{subfloat}[]{
			\includegraphics[width=0.14\textwidth]{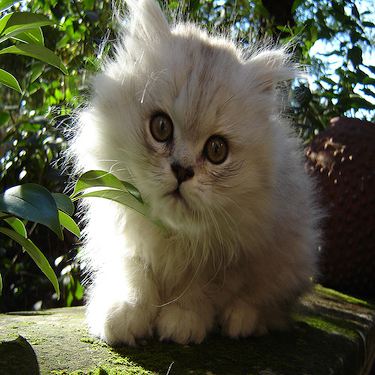}
			\label{fig:resnet_feature_map_orig}
		}
	\end{subfloat}
	\begin{subfloat}[]{
			\includegraphics[width=0.14\textwidth]{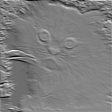}
			\label{fig:resnet_feature_map_1}
		}
	\end{subfloat}
	\begin{subfloat}[]{
			\includegraphics[width=0.14\textwidth]{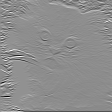}
			\label{fig:resnet_feature_map_2}
		}
	\end{subfloat}
	
	\caption{Sample feature maps of ResNet. (a) shows the original image of a Persian cat. (b) and (c) are feature maps extracted from ResNet. }
	\label{fig:resnet_feature_maps}
\end{figure}

It can be observed from the ResNet architecture that the only output is the prediction of the input image's label, and the whole ResNet is a black-box. The layers between the input and the output are adopted to construct feature maps. In the ResNet drawn in Fig. \ref{fig:cnn-blocks-resnet}, 64 feature maps, $4\times64$ features maps and $6\times128$ feature maps are constructed by the first blue convolutional layer, the group of four orange convolutional layers, and the group of 6 purple convolutional layers, respectively. In total, there are thousands of feature maps constructed by ResNet. The visualised feature maps are shown in Fig.  \ref{fig:resnet_feature_maps}. Fig. \ref{fig:resnet_feature_map_1} and Fig.  \ref{fig:resnet_feature_map_2} are two sample feature maps learned by ResNet from a Persian cat of Fig. \ref{fig:resnet_feature_map_orig}. The contour of the Persian cat could be seen in the two feature maps, but the two contour images look similar. It is hard to tell why ResNet generates two similar feature maps, and it is also unlikely that humans could tell what kind of cat the two feature maps indicate. Overall, ResNet is a black-box because inside ResNet, there are thousands of non-intuitive feature maps constructed that lead ResNet to make the prediction, but these feature maps do not make much sense to humans. 

\begin{figure*}[ht]
	\centering
	\includegraphics[width=0.8\linewidth]{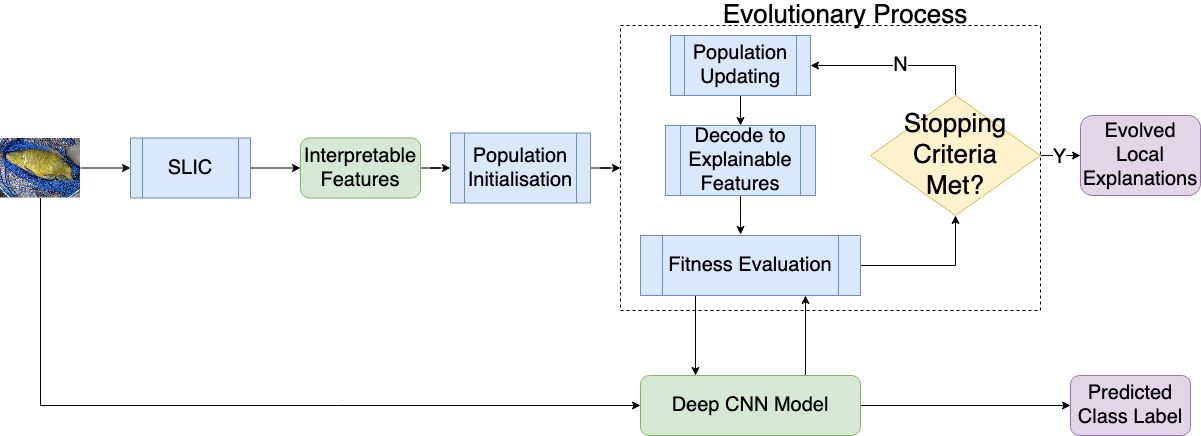}
	\caption{Overall framework.}
	\label{fig:elime_framework}
\end{figure*}

\subsection{Explainable Deep Learning Methods}

Explainable deep learning (XDL) can be divided into three categories - \textit{Visualisation methods}, \textit{Intrinsic methods} and \textit{Model distillation} \cite{xie2020explainable}. Visualisation methods are a popular class of XDL methods, which learn heat maps or saliency maps to highlight the important features of input images. Examples of such methods include Deconvolution \cite{zeiler2014visualizing}, CAM \cite{zhou2016learning}, and DeepLIFT, \cite{shrikumar2017learning}. Most visualisation methods add visualisation layers, such as DeConv layers in Deconvolution \cite{zeiler2014visualizing}, to deep convolutional neural networks (deep CNNs) to be explained. The modification of deep CNNs requires deep learning expertise. In addition, the modified deep CNNs with visualisation layers often need to be re-trained, which incurs relatively high computational costs.

Intrinsic methods simultaneously provide the decisions and their corresponding explanations as the output. Single-Modal Weighting \cite{devlin2018bert}, Explanation Association \cite{iyer2018transparency} and Model Prototype \cite{chen2019looks} are some examples of intrinsic methods. Similar to visualisation methods, intrinsic methods require expert knowledge to re-design deep CNNs to be explained, and computing resources to re-train the modified deep CNNs. For example, Prototype \cite{chen2019looks} designs prototype layers which are added to deep CNNs to be explained. The re-designed deep CNNs are called ProtoNet, which need to be re-trained.

Model distillation methods distil the knowledge encoded in DCNNs into a more interpretable representation. One branch of this method is model translation, which is to learn a simpler model to replicate the behaviours of DCNNs \cite{frosst2017distilling} \cite{zhang2019interpreting}. However, this may sacrifice performance and require domain expertise to understand the behaviours of the simple model. The other branch is local approximation, with LIME \cite{ribeiro2016should} being the most popular method. LIME finds a subset of interpretable features that are easy to be understood by end-users without requiring machine learning expertise. However, it suffers from the issue of high computational cost and requires pre-defining the number of interpretable features.

\subsection{Related Work}
Most existing works attempt to explain the predictions of a complex model as a whole. However, it is usually complex to explain a black-box model globally, but is relatively easy to explain each of the predictions made by the model. In this study, we focus mainly on learning local explanations to describe the behaviours of deep CNN models. Recently, most popular methods for learning local explanations are LIME \cite{ribeiro2016should} and anchors \cite{ribeiro2018anchors}. 

LIME proposed by Ribeiro et al. \cite{ribeiro2016should} is targeted at explaining predictions of a black-box classifier or regressor in an interpretable or a sensible way, and thereby provides humans with insights into the black-box model. An interpretable model (i.e., an explanation) is defined as a linear model, or decision trees, etc. The main idea here is that an interpretable model is used to approximate the black-box model locally. In more detail, there are three main steps in LIME to explain a prediction made by the black-box model for an unseen instance $x'$ as follows. Firstly, the instances around $x'$ are randomly sampled and weighed based on its distance to $x'$. Secondly, the black-box model is used to predict labels (or probabilities) of the weighted instances. Thirdly, these weighed instances with the predicted values are used to train a simple model (i.e., an interpretable model). 

However, LIME does not consider the neighbourhood of instance $x'$. Therefore, it is unclear whether an explanation for $x'$ can be used for its neighbours, i.e., its coverage region (where the explanation applies) is not clear. In LIME, a complex model is approximated locally by a linear model, but it is likely that a complex model exhibits non-linear behaviours in the neighbourhood of instance $x'$. Therefore, LIME explanations may be misleading when explaining predictions on similar unseen instances. To address this drawback, Ribeiro et al. \cite{ribeiro2018anchors} introduced another model-agnostic explanations method (named \textit{anchors}) to explain the behaviours of complex models by maximising the coverage region of an explanation. An anchor explanation is an if-then rule, which is easy to understand. However, \textit{anchors} may not work well for predictions that are near a boundary of the black box model's decision function, or predictions of very rare classes \cite{ribeiro2018anchors}. 

The use of EC techniques for learning explanations has not been investigated for image classification. In this paper, we attempt to use a binary GA to automatically learn explanations to approximate a complex CNN model locally, in order to explain its behaviours in an understandable way. As introduced previously, in LIME, humans need to define explanations, such as decision trees, to be trained for explaining a complex model. The use of GAs for learning explanations is independent of human expertise.

\section{The Proposed Method} \label{SSS:elime_method}

In this section, we introduce the overall framework of the proposed method first, and then describe each component in more detail.  
\subsection{Overall Framework}

The overall framework of the proposed method, E-LIME, is illustrated in Fig.  \ref{fig:elime_framework}. The framework consists of two main pathways: (1) the generation of local explanations using Genetic Algorithm (GA), and (2) the prediction of the input image using a pre-trained deep CNN model.

\begin{itemize}
    \item \textbf{Superpixel Generation}: The input image is first processed using the Simple Linear Iterative Clustering (SLIC) algorithm \cite{achanta2012slic}, which segments the image into superpixels. These superpixels serve as interpretable features that can be easily understood by humans.
    \item \textbf{Population Initialization}: The initial population of GA is constructed by randomly generating binary vectors, where each vector represents a candidate solution. The dimensionality of the binary vectors corresponds to the number of superpixels. A value of 1 indicates that the corresponding superpixel is selected, while a value of 0 indicates that it is not selected.
    \item \textbf{Evolutionary Process}: Selection, crossover, and mutation are applied to evolve the population. Each individual in the population is decoded into a set of selected superpixels, forming a local explanation. The fitness of each individual is evaluated based on the probability of the deep CNN model predicting the correct class label for the masked image (i.e., the image with only the selected superpixels).
    \item \textbf{Final Explanation}: The best individual from the final generation is decoded into the evolved local explanation, which highlights the most important superpixels for the prediction.
\end{itemize}

On the bottom pathway, the pre-trained deep CNN model performs a prediction on the input image. The predicted class label, along with the evolved local explanation, is presented to the end-user. If the interpretable features in the local explanation align with the characteristics of the predicted class, the end-user can have greater confidence in the prediction.

\begin{figure}[ht]
	\centering
	\begin{subfloat}[The original image.]{
		\includegraphics[width=0.22\textwidth]{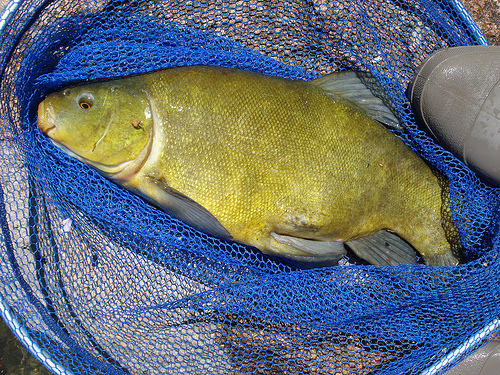}
		\label{fig:original_image}
	}
	\end{subfloat}
	\begin{subfloat}[A sample uninterpretable feature map of ResNet. ]{
		\includegraphics[width=0.22\textwidth, height=0.125\textheight]{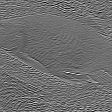}
		\label{fig:unexplainable_feature_map_2}
	}
	\end{subfloat}
	\begin{subfloat}[Superpixels, i.e., segments obtained by SLIC. ]{
		\includegraphics[width=0.22\textwidth]{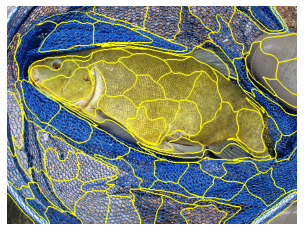}
		\label{fig:image_superpixels}
	}
	\end{subfloat}
	\begin{subfloat}[Evolved interpretable features. ]{
		\includegraphics[width=0.22\textwidth]{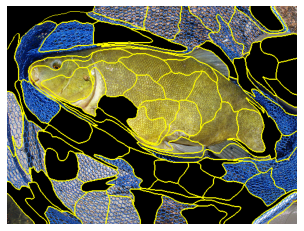}
		\label{fig:exp_features}
	}
	\end{subfloat}
	
	\caption{Interpretable features vs uninterpretable feature maps. }
	\label{fig:exp_features_vs_unex_features}
\end{figure}

\subsection{SLIC to Generate Interpretable Features}

Deep CNNs use convolutional layers to perform feature extractions, but the extracted features are hard to be interpreted by humans. This is also one of the reasons why deep CNNs are considered to be black-box models. The convolutional layers extract features from the input feature maps to output feature maps. The original image can be deemed as a single feature map for greyscale images and three feature maps for RGB colour images. Fig.  \ref{fig:unexplainable_feature_map_2} shows an example of one of the feature maps extracted by the first convolutional layer of ResNet\cite{he2016deep}. The texture of a tench fish can be vaguely noticed by humans, but the texture could be any fish. It gets worse when the feature maps are extracted from deeper layers, which could not be recognised by humans at all. Therefore, the features extracted by deep CNNs are not suitable to show how deep CNNs extract features to make the predictions. These features are non-interpretable features. 

To explain deep CNNs, the superpixels are adopted as the interpretable features,  inspired by LIME \cite{ribeiro2016should}. SLIC is used to generate the superpixels. Fig. \ref{fig:image_superpixels} exhibits 100 superpixels generated by SLIC from the original image shown in Fig. \ref{fig:original_image}. As can be observed, each superpixel contains a set of similar pixels nearby, which can be easily distinguished by humans. This is why the superpixels can be used as interpretable features. 

The main goal of the proposed method is to select a subset of important features from all of the interpretable features generated by SLIC from the original image, which can explain how deep CNNs make their predictions. Fig.  \ref{fig:exp_features} presents an example of the selected interpretable features as the local explanation of a deep CNN model. The suerpixels, i.e., the segments in the black colour are the features that are not selected, while the other superpixels with the original colours are the selected features. The selected interpretable features make sense for predicting the input image as a tench fish due to the following observations. Firstly, as an important characteristic of a tench fish, the interpretable feature containing the brick red eye is selected. The deep CNN model  makes the prediction by taking this distinct feature of a tench fish into account, which aligns with human knowledge. Therefore, it may improve the confidence of real users. Secondly, most parts of the fish body are selected as important features. This also improves the trust of the model prediction because the body parts are crucial for distinguishing a fish. But the superpixels with the front fin of the fish are not selected. This could mean that this front fin may not be a good feature to distinguish a tench from other fish. 

\subsection{Encoding Strategy}\label{sec:elime_encoding}

\begin{figure}[ht]
	\centering
	\includegraphics[width=0.6\linewidth]{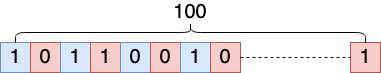}
	\caption{Encoding strategy.}
	\label{fig:elime_encoding}
\end{figure}

To encode the interpretable features of a local explanation, a binary vector shown in Fig.  \ref{fig:elime_encoding} is used to encode the corresponding features. The number of superpixels $n_s$ needs to be defined, which is a hyper-parameter of the proposed method. The target image is then segmented into $n_s$ superpixels, i.e., $n_s$ interpretable features. In the example shown in Fig.  \ref{fig:elime_encoding}, $n_s$ is set to 100, which will generate 100 superpixels (interpretable features), the same as  Fig. \ref{fig:image_superpixels}. To encode those 100 features, they are marked as $1st$ to $100th$ features from the top-left to the bottom-right. Each dimension in the encoded vector indicates whether the corresponding feature is selected or not, i.e., 1 means the corresponding feature is selected, and 0 otherwise. In Fig. \ref{fig:elime_encoding}, the $1st$ dimension is $1$ and the $2nd$ dimension is $0$. This means that the $1st$ interpretable feature is selected, while the $2nd$ is not selected. 

\subsection{Decoding a Vector to a Local Explanation} \label{sec:elime_decoding}

The encoded vector needs to be decoded to an image similar to Fig.  \ref{fig:exp_features}. The $0s$ in the encoded vectors are converted to the black superpixels, and the $1s$ means the interpretable features are kept in the decoded image. There are two steps that require the decoding process. The first is the fitness evaluation process, where the decoded image is taken as the input of deep CNNs. The second is to return the selected interpretable features---the last box in the first row of Fig. \ref{fig:elime_framework}. The decoded image is recognisable for humans, and humans can tell whether the selected interpretable features are reliable for deep CNNs to make the prediction in order to make further decisions on trusting the predictions or not. 

\subsection{Fitness Evaluation}

A softmax function \cite{goodfellow2016deep} defined in Formula (\ref{eqn:softmax}) is appended to the last linear layer of the pre-trained deep CNN model to obtain a probability distribution over the $K$ possible class labels. 

\begin{equation}\label{eqn:softmax}
\sigma (\overrightarrow{z})_{i}=\frac{e^{z_{i}}}{\sum_{j=1}^{K}e^{z_{j}}}
\end{equation}
where $\sigma (\overrightarrow{z})_{i}$ indicates the probability of the image belonging to the $i_{th}$ class label, $\overrightarrow{z}$ is the input vector made up of $(z_{1}, z_{2}, ..., z_{K})$ to the softmax function, and $z_{i}$ represents the $i_{th}$ element in the input vector.  Suppose the predicted class for an input image by the pre-trained deep CNN model is the class label $i$, the corresponding probability is calculated based on the above formula, which is used as the fitness value. 

\begin{algorithm}[h]
	\caption{Fitness evaluation}
	\label{alg:elime_fitness_evaluation}
	\begin{algorithmic}[1]
		\renewcommand{\algorithmicrequire}{\textbf{Input:}}
		\renewcommand{\algorithmicensure}{\textbf{Output:}}
		\newcommand{\algorithmicbreak}{\textbf{break}}
		\newcommand{\BREAK}{\STATE \algorithmicbreak}
		\REQUIRE pre-trained deep CNN model $model$, original image $img_{o}$, individual $ind$ ;
		\STATE Use $model$ to predict the class label, $label$, for $img_{o}$;
		\STATE Decode $ind$ to produce a masked image, $img_{m}$, from $img_{o}$ according to Section \ref{sec:elime_decoding};
		\STATE Use $model$ to perform a prediction of $img_m$ and get the probability, $prob$, of $label$;
		\STATE $fitness \leftarrow prob$; 
		\RETURN $fitness$;
	\end{algorithmic}
\end{algorithm}

Algorithm \ref{alg:elime_fitness_evaluation} describes the details of the fitness evaluation function. Since the proposed method is model-agnostic, any pre-trained deep CNN model to be explained can be passed to the fitness evaluation step. As a local explanation method, the original image that needs to be locally explained is also required. The above two characteristics define the proposed method as a \textit{local interpretable model-agnostic} method. The actual evaluation process is straightforward mainly including three steps. The first step is to predict the class label of the original image by the pre-trained model. The second is to decode the GA individual to a local explanation according to the decoding process described in Section \ref{sec:elime_decoding}. The local explanation is a masked image, which consists of the interpretable features represented by the GA individual. In the end, the pre-trained model performs a prediction on the masked image, and the probability of having the same class label as the original image is obtained, which is then used as the fitness value of the individual. The target of the proposed method is to optimise the fitness value, i.e., the probability of the class label of the original image. The optimisation process can extract the features that are important for the deep CNN model to make the prediction, and remove noisy features that are misleading for the learning process.

\subsection{Evolving Local Explanations}

\begin{algorithm}[h]
	\caption{Evolve local explanations}
	\label{alg:elime_evolution}
	\begin{algorithmic}[1]
		\renewcommand{\algorithmicrequire}{\textbf{Input:}}
		\renewcommand{\algorithmicensure}{\textbf{Output:}}
		\newcommand{\algorithmicbreak}{\textbf{break}}
		\newcommand{\BREAK}{\STATE \algorithmicbreak}
		\REQUIRE the population size $s$, the number of generations $g_{max}$, a local image $img$;
		\STATE $pop \leftarrow$ Randomly initialise the population with $s$ individuals;
		\STATE Evaluate every individual in $pop$ using the fitness evaluation illustrated in Algorithm \ref{alg:elime_fitness_evaluation} based on $img$;
		\STATE $best \leftarrow$ Select the best individual from $pop$;
		\STATE $g_{cur} \leftarrow 0$;
		\WHILE{$g_{cur} < g_{max}$}
		\STATE Apply tournament selection to select parents from $pop$;
		\STATE Apply the crossover operator; 
		\STATE Apply the mutation operator to generate new individuals to form $offspring$;
		\STATE Apply fitness evaluation as described in Algorithm \ref{alg:elime_fitness_evaluation} based on $img$ to evaluate individuals in $offspring$;
		\STATE $pop \leftarrow offspring$;
		\STATE $g_{best} \leftarrow $ Select the best individual from $pop$;
		\IF{$g_{best} > best $}
		\STATE $best \leftarrow g_{best}$;
		\ENDIF
		\ENDWHILE
		\RETURN $best$;
	\end{algorithmic}
\end{algorithm}

After the encoding strategy and the evaluation process are designed, it is straightforward to start the GA process to evolve the local explanations. During the evolutionary learning process, a local explanation is evolved for a local image instance. A local explanation is to explain why the pre-trained deep CNN model predicts a specific class label for a specific input image, where the specific input image is the local image. The main steps in the evolutionary process are introduced as follows. Firstly, a population is randomly initialised. Secondly, the fitness function is utilised to evaluate every individual in the population, and the best individual is selected and saved. Based on the fitness values, the good individuals are selected by the tournament selection, and the selected individuals are fed to the  mutation and crossover operations to generate new offspring for the next population. Thirdly, every individual in the new population is evaluated, and the best individual is updated if the best individual is better than the current best individual. In the end, the best individual is returned, which is then decoded to a masked image as the explanation of the deep CNN on the local image. 

\subsection{Computational Complexity Analysis}

The computational complexities of E-LIME and LIME are analysed to provide theoretical support for the efficiency of the proposed method. LIME's complexity is dominated by the sampling process and linear model training, resulting in a complexity of:
\[
O(N \cdot T_{\text{CNN}} + N \cdot n_s^2),
\]
where \( N \) is the number of perturbed instances, \( T_{\text{CNN}} \) is the time required for a single forward pass through the deep CNN model, and \( n_s \) is the number of superpixels.

In contrast, E-LIME eliminates the sampling process and uses a genetic algorithm (GA) to optimise the selection of superpixels, resulting in a complexity of:
\[
O(P \cdot G \cdot T_{\text{CNN}} + P \cdot n_s),
\]
where \( P \) is the population size and \( G \) is the number of generations. This analysis demonstrates that E-LIME is more computationally efficient than LIME, especially for large \( n_s \). The GA in E-LIME typically converges in a small number of generations (e.g., 20-30), further enhancing its efficiency.

\section{Experiment Design} \label{SSS:elime_experiment}

\subsection{Benchmark Dataset}

The ImageNet dataset \cite{krizhevsky2012imagenet} is selected as the benchmark dataset to evaluate the proposed method. This is mainly because the ImageNet is comprised of real-world images collected from flickr and other search engines, and the proposed method is expected to explain the predictions of deep CNNs on real-world images rather than just working for synthetic or preprocessed datasets, such as the CIFAR \cite{krizhevsky2009learning} or SVHN \cite{netzer2011reading} datasets.   There are 1,000 class labels in the ImageNet dataset. The training set consists of 1.2 million images and the validation set contains 150,000 images. 

The deep CNN models to be explained, are trained on the training set of the benchmark dataset. The trained model then predicts the class labels of some images from the test set. As the proposed method is a local explanation method (i.e., the local explanation evolved by the proposed method is for a specific image), it is not possible to list all of the local explanations for the 150,000 images from the validation set. In this paper, due to the page limit, only the local explanations of 4 images, which are randomly chosen from the validation set, will be displayed and explained. 

\subsection{Selected Deep Neural Networks}

ResNet \cite{he2016deep}, which is one of the state-of-the-art deep CNN models in recent years, is adopted as the main deep CNN model to be explained. A couple of major considerations are that ResNet is well-known in the deep learning community, and the pre-trained ResNet is available in many deep learning frameworks, such as Pytorch and Tensorflow. This also makes the replication of this research fairly easy for researchers with different backgrounds. 

However, the proposed method is model-agnostic due to the benefits of the fitness evaluation designed in Algorithm \ref{alg:elime_fitness_evaluation}. ResNet could be replaced by any deep CNN model, e.g. VGGNet \cite{simonyan2014very}, Xception \cite{chollet2017xception}, DenseNet \cite{huang2017densely}, or MobileNet \cite{howard2017mobilenets}, but the deep CNN model needs to be trained on the training set first, which can then be taken as the pre-trained model in the fitness evaluation. 

Therefore, to further evaluate the model-agnostic characteristic of the proposed method, two other deep CNN models --- DenseNet \cite{huang2017densely} and MobileNet \cite{howard2017mobilenets} are selected as two additional representative deep learning models to be explained. The major reasons of choosing DenseNet and MobileNet are that their architectures are considerably different from ResNet, and they are also as popular as ResNet besides the easy accessibility of their pre-trained models. 

\begin{figure}[ht]
	\centering
	\includegraphics[width=\linewidth]{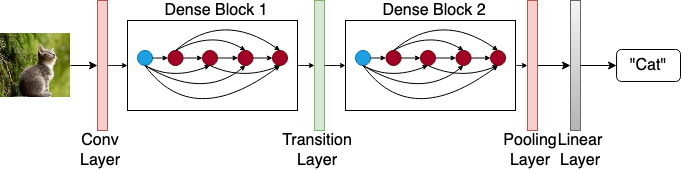}
	\caption{DenseNet architecture \cite{huang2017densely}.}
	\label{fig:cnn-blocks-densenet}
\end{figure}

Fig. \ref{fig:cnn-blocks-densenet} displays a DenseNet architecture with two dense blocks. The input image is processed by a convolutional layer to produce a set of feature maps. The feature maps are passed to the first dense block to construct a new set of feature maps. A transition layer, which consists of a convolutional layer and a pooling layer, is utilised to connect the two dense blocks. At the end, a pooling layer and a linear layer take the output feature maps of the second dense block to predict the class label of the input image. With regard to the general network architecture, there are two obvious differences between DenseNet and ResNet. Firstly, the whole network of DenseNet splits the architecture into several dense blocks instead of having one big block in the ResNet architecture. Furthermore, the shortcut connections in dense blocks connects the current layer to all of the latter layers, while in ResNet the shortcut connections are only added to the current layer and the layer after its next layer. As a result, DenseNet could represent another type of deep CNN models that is significantly different from ResNet. 

\begin{figure}[ht]
	\centering
	\includegraphics[width=\linewidth]{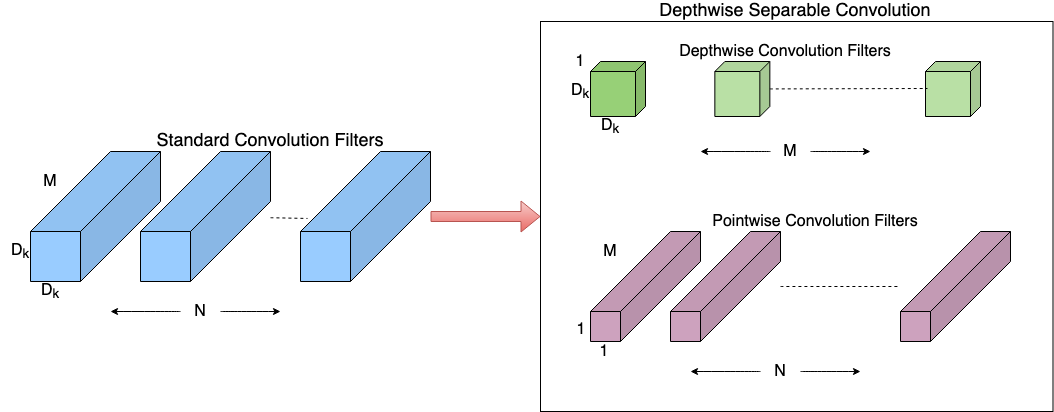}
	\caption{The transformation from the standard convolution layer to the depthwise separable convolutional layer used in MobileNet \cite{howard2017mobilenets}.}
	\label{fig:cnn-blocks-mobilenet}
\end{figure}

MobileNet has a fundamental modification of the standard convolutional layer used in ResNet and DenseNet. Fig. \ref{fig:cnn-blocks-mobilenet} illustrates the transformation from the standard convolutional layer to the so-called \textit{depthwise separable convolutional layer} \cite{howard2017mobilenets}. $D_{k} \times D_{k}$ represents the filter size (width $\times$ height), $M$ is the number of channels of the image, and $N$ is the number of the output feature maps. MobileNet splits the standard convolution filters into two separable convolutional filters --- the \textit{depthwise convolution filters} and the \textit{pointwise convolution filters}. Given the image size of $D_{f} \times D_{f}$, the computational cost of the standard convolution filters is $D_{k} \times D_{k} \times M \times N \times D_{f} \times D_{f}$, while the computational cost of the depthwise separable convolution filters is $D_{k} \times D_{k} \times M \times D_{f} \times D_{f} + M \times N \times D_{f} \times D_{f}$. It can be observed that the depthwise separable convolution filters require less computational resource than the standard ones. This is why MobileNet replaces the standard convolution filters with the depthwise separable convolution filters. 

In terms of the specific models used in the experiments, 
ResNet-34 \cite{he2016deep}, DenseNet-121 \cite{huang2017densely}, and MobileNetV2 \cite{sandler2018mobilenetv2} are selected to represent different types of CNN models, which are used to evaluate the explainability and the model-agnostic characteristic of the proposed method. 

\subsection{Parameter Settings}

\begin{figure*}[!ht]
	\centering
	\begin{subfloat}[]{
			\includegraphics[width=0.22\textwidth]{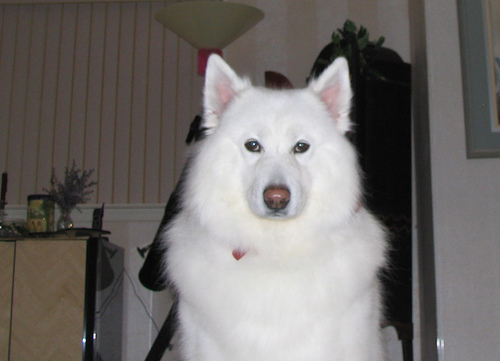}
			\label{fig:elime_dog_orig}
		}
	\end{subfloat}
	\begin{subfloat}[]{
			\includegraphics[width=0.22\textwidth]{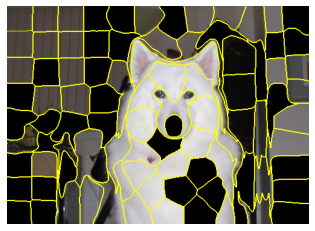}
			\label{fig:elime_dog_resnet}
		}
	\end{subfloat}
	\begin{subfloat}[]{
			\includegraphics[width=0.22\textwidth]{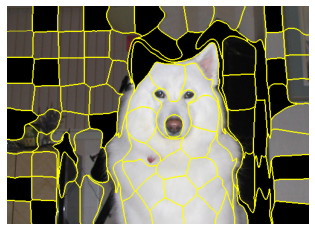}
			\label{fig:elime_dog_densenet}
		}
	\end{subfloat}
	\begin{subfloat}[]{
			\includegraphics[width=0.22\textwidth]{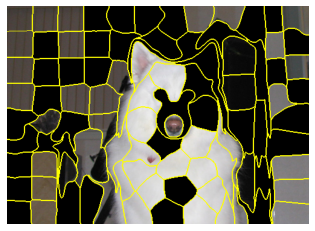}
			\label{fig:elime_dog_mobilenet}
		}
	\end{subfloat}
	\caption{Evolved local explanations for the image of a Samoyed. From left to right, it is the original image, evolved local explanations based on \textbf{ResNet}, \textbf{DenseNet} and \textbf{MobileNet}, respectively.}
	\label{fig:elime_dog}
\end{figure*}

\begin{figure*}[!ht]
	\centering
	\begin{subfloat}[]{
			\includegraphics[width=0.22\textwidth]{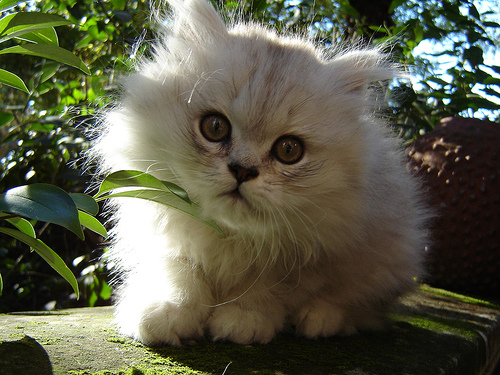}
			\label{fig:elime_cat_orig}
		}
	\end{subfloat}
	\begin{subfloat}[]{
			\includegraphics[width=0.22\textwidth]{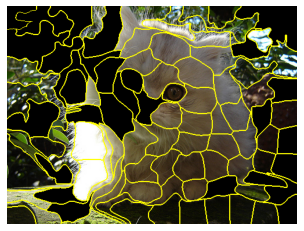}
			\label{fig:elime_cat_resnet}
		}
	\end{subfloat}
	\begin{subfloat}[]{
			\includegraphics[width=0.22\textwidth]{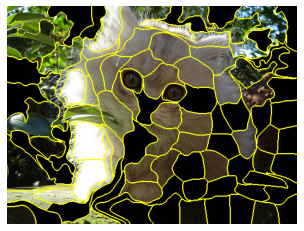}
			\label{fig:elime_cat_densenet}
		}
	\end{subfloat}
	\begin{subfloat}[]{
			\includegraphics[width=0.22\textwidth]{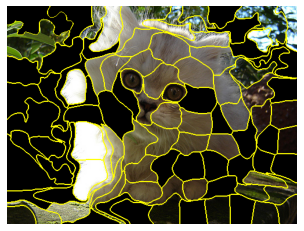}
			\label{fig:elime_cat_mobilenet}
		}
	\end{subfloat}
	\caption{Evolved local explanations for the image of a Persian cat. From left to right, it is the original image, evolved local explanations based on \textbf{ResNet}, \textbf{DenseNet} and \textbf{MobileNet}, respectively.}
	\label{fig:elime_cat}
\end{figure*}


There are two sets of parameters in the proposed method, i.e., the parameters of GAs and the hyper-parameters specific to the proposed method. The GA parameters are set according to the EC community convention recommended in \cite{back1996evolutionary}. Specifically, the crossover rate, mutation rate, population size, and number of generations are set to 0.9, 0.2, 100 and 50, respectively. There is only one hyper-parameter specific to the proposed method, which is the number of superpixels $n_s$, i.e., the number of interpretable features in this paper.  There are two factors that determine the value of $n_s$. The first one is the ratio of the target object in the whole image. For example, if the target object only occupies $1/10$ of the whole image, $n_s$ should be significantly larger than $10$. Otherwise, even though the superpixel that only contains the target object is selected, there are still some noisy background pixels included in the selected feature. Another factor is the maximum time expected to produce the explanations. Since a larger value $n_s$ could result in a larger search space, the proposed method may require more time to converge to obtain the evolved explanations. By taking the above two factors into account, $100$ is set as the value of $n_s$ in our experiments on the ImageNet dataset. Specifically, we examined the values of 20, 50, 100, 150 and 200 as $n_s$, 100 is the smallest one that can separate object features from the background. All the experiments were ran on a single GPU card with the specific model of GeForce GTX 1080. 

\section{Results and Analysis}\label{SSS:elime_result}

In the experiments, the proposed method is used to process four images\footnote{The local explanations of 100 more images are provided here: https://github.com/wwwbbb8510/elime-report.} randomly selected from the ImageNet dataset to obtain the local explanations. 

\subsection{Evolved Local Explanations}\label{SSS:elime_result_evolved_explanations}
\subsection*{1) Evolved explanations on the image of a Samoyed}
Fig.  \ref{fig:elime_dog_orig} shows the original image of a Samoyed. In Fig. \ref{fig:elime_dog_orig}, several characteristics of the Samoyed can be observed -- a white and dense coat, a pair of upright and prick ears, and brown eyes and nose. In particular, a red collar can be seen on its neck. Besides, some background objects exhibit -- a cabinet, a box and a vase of flowers above the cabinet, and a hanger behind the Samoyed.

Fig. \ref{fig:elime_dog_resnet} shows the evolved local explanation, which attempts to assist humans to understand the prediction of ResNet on the image of a Samoyed, by using E-LIME. As can be seen from Fig. \ref{fig:elime_dog_resnet}, the shape and size of ears are extracted in the local explanation, which are the most informative features. The features of upright and prick ears, and the size of each ear differentiate Samoyeds from others. Apart from that, the informative feature related to eyes is also extracted in the local explanation. After knowing these extracted explanations, users may feel confident to trust in the prediction. In addition, most of the background objects, which could become noisy features, are filtered out, e.g. a vase of flowers and the hanger. This could further improve the confidence in the prediction and provide evidence for users to believe the prediction. Accordingly, after presenting the prediction as well as its local explanation to users, it is relatively easier for end-users to decide whether to trust the predicted class label or not.

Fig. \ref{fig:elime_dog_densenet} and Fig. \ref{fig:elime_dog_mobilenet} show the explanations evolved by E-LIME to explain the predictions of DenseNet and MobileNet, respectively, on the same image. In Fig. \ref{fig:elime_dog_densenet}, important features are also effectively extracted as the local explanation, such as the eyes and nose. Slightly different from that in Fig. \ref{fig:elime_dog_resnet} for ResNet, only the right ear is extracted in the local explanation. However, after considering other extracted interpretable features as supplemental materials (such as nose and eyes), it is sufficient for humans to understand the prediction although only the right ear is extracted in the local explanation (the left ear may become a redundant feature if other effective features are extracted). The similar situation also exists in Fig. \ref{fig:elime_dog_mobilenet}. The left ear and the nose are effectively extracted in the local explanation, which are sufficient for MobileNet to make a prediction without other information about eyes and the right ear. In addition, in Fig. \ref{fig:elime_dog_densenet} and Fig. \ref{fig:elime_dog_mobilenet}, most irrelevant information is filtered in order to effectively distinguish between the Samoyed and the background information. 

It is experimentally proved that the proposed method is model-agnostic, i.e., it is applicable to extract local explanations for explaining predictions or behaviours of any deep CNN model. This is mainly because the proposed fitness function is able to evaluate the selected superpixels (i.e., interpretable features) represented by the encoded vector based on any deep CNN model.

\subsection*{2) Evolved explanations on the image of a Persian cat}

In Fig. \ref{fig:elime_cat_orig}, a white Persian cat stands on the ground with some moss, under the shadow of trees. Fig. \ref{fig:elime_cat_resnet}, Fig. \ref{fig:elime_cat_densenet} and Fig. \ref{fig:elime_cat_mobilenet} show the learned local explanations for the image of a Persian cat by using E-LIME to explain the predictions of ResNet, DenseNet and MobileNet, respectively. 

As can be seen from Fig. \ref{fig:elime_cat_mobilenet}, useful and informative interpretable features, such as ears, eyes, nose and mouth, are learned to intuitively describe features of the cat. This is able to explain why the prediction is a cat to some extent. For example, the difference between the cat and the dog (shown in Fig. \ref{fig:elime_dog}) is relatively obvious when comparing their shapes of eyes, nose and mouth. In Fig. \ref{fig:elime_cat_resnet}, information related to the left eye, nose and mouth is not extracted in the local explanation for ResNet. This is mainly because the extracted interpretable features are enough for ResNet to make a prediction and also relatively sufficient for users to understand the prediction. Similarly, in Fig. \ref{fig:elime_cat_densenet}, nose and mouth are not extracted in the local explanation based on DenseNet. In addition, the extracted interpretable features could further distinguish the silhouette of the cat from the background of this image, such as the leaves and ground. In Fig. \ref{fig:elime_cat_orig}, it is noticed that two of the green leaves touch the left lower jaw of the cat. As can be shown from Fig. \ref{fig:elime_cat_resnet} and Fig. \ref{fig:elime_cat_mobilenet}, the two leaves are discerned and not selected because these two leaves are not helpful to make the prediction of a cat. This enhances the chance of predicting this image into the cat category. 
Therefore, E-LIME does provide intuitive explanations to users who want to know the reasons behind the prediction in order to determine whether the prediction is reasonable and trustworthy or not. 

Therefore, E-LIME does provide intuitive explanations to users who want to know the reasons behind the prediction in order to determine whether the prediction is reasonable and trustworthy or not. 

\subsection*{3) Evolved explanations on the image of a bald eagle}

\begin{figure*}[!ht]
	\centering
	\begin{subfloat}[]{
			\includegraphics[width=0.22\textwidth]{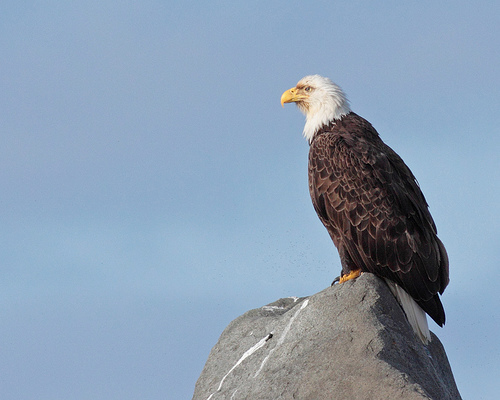}
			\label{fig:elime_eagle_orig}
		}
	\end{subfloat}
	\begin{subfloat}[]{
			\includegraphics[width=0.22\textwidth]{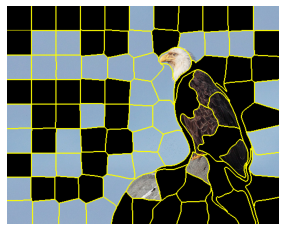}
			\label{fig:elime_eagle_resnet}
		}
	\end{subfloat}
	\begin{subfloat}[]{
			\includegraphics[width=0.22\textwidth]{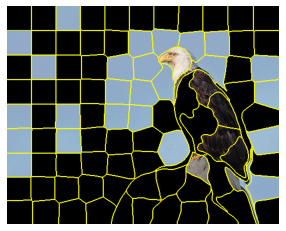}
			\label{fig:elime_eagle_densenet}
		}
	\end{subfloat}
	\begin{subfloat}[]{
			\includegraphics[width=0.22\textwidth]{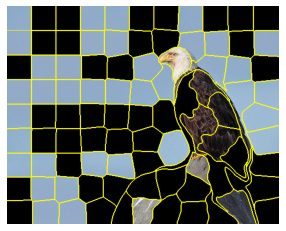}
			\label{fig:elime_eagle_mobilenet}
		}
	\end{subfloat}
	\caption{Evolved local explanations for the image of a bald eagle. From left to right, it is the original image, evolved local explanations based on \textbf{ResNet}, \textbf{DenseNet} and \textbf{MobileNet}, respectively.}
	\label{fig:elime_eagle}
\end{figure*}

\begin{figure*}[!ht]
	\centering
	\begin{subfloat}[]{
			\includegraphics[width=0.22\textwidth]{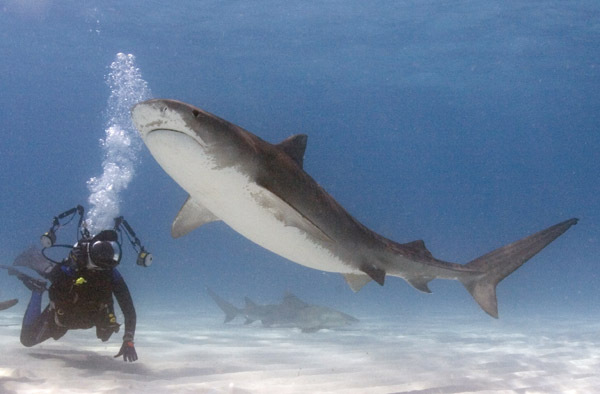}
			\label{fig:elime_shark_orig}
		}
	\end{subfloat}
	\begin{subfloat}[]{
			\includegraphics[width=0.22\textwidth]{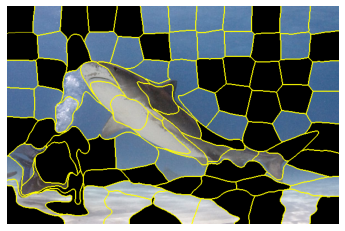}
			\label{fig:elime_shark_resnet}
		}
	\end{subfloat}
	\begin{subfloat}[]{
			\includegraphics[width=0.22\textwidth]{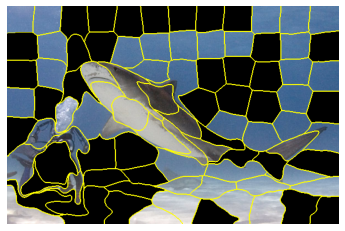}
			\label{fig:elime_shark_densenet}
		}
	\end{subfloat}
	\begin{subfloat}[]{
			\includegraphics[width=0.22\textwidth]{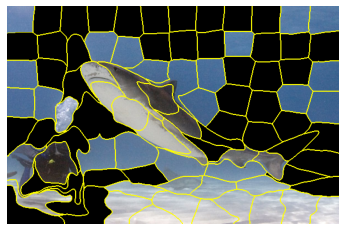}
			\label{fig:elime_shark_mobilenet}
		}
	\end{subfloat}
		\caption{Evolved local explanations for the image of a tiger shark. From left to right, it is the original image, evolved local explanations based on \textbf{ResNet}, \textbf{DenseNet} and \textbf{MobileNet}, respectively.}
	\label{fig:elime_shark}
\end{figure*}

Fig. \ref{fig:elime_eagle_orig} shows a bald eagle standing on a rock. The eagle in this image has a brown body and brown wings (drooped in this image), a white head with a hooked yellow beak, and yellow feet. 

Based on Fig. \ref{fig:elime_eagle_resnet}, the contour of the head with a hooked beak is learned and extracted in the local explanation to interpret the prediction of ResNet on the image. Similarly, the useful explanation can be also found in Fig. \ref{fig:elime_eagle_densenet} and Fig. \ref{fig:elime_eagle_mobilenet} for DenseNet and MobileNet, respectively. In fact, the shape of the hooked beak is one of the most important features to effectively recognise an eagle from an image. Therefore, this extracted explanation could show why the prediction is an eagle. Besides, as shown in Fig. \ref{fig:elime_eagle_resnet}, Fig. \ref{fig:elime_eagle_densenet} and Fig. \ref{fig:elime_eagle_mobilenet}, information related to the body silhouette of the eagle is also extracted in the local explanation, which are able to discern the difference between the eagle and the rock. 
Moreover, the irrelevant features about the rock are also effectively removed in Fig. \ref{fig:elime_eagle_resnet}, Fig. \ref{fig:elime_eagle_densenet} and Fig. \ref{fig:elime_eagle_mobilenet}. Therefore, after showing these local explanations to users, they may find it easier to understand the prediction and become confident to trust it. 

\subsection*{4) Evolved explanations on the image of a tiger shark}

In Fig. \ref{fig:elime_shark_orig}, a tiger shark is swimming in the ocean, and on the lower left corner, there is a diver who exhales some bubbles. The shark has pectoral and dorsal fins, and a crescent-shaped tail. The belly of the shark is whitish, while its back and sides are gray. This image is predicted to the shark category. Fig. \ref{fig:elime_shark_resnet}, Fig. \ref{fig:elime_shark_densenet} and Fig. \ref{fig:elime_shark_mobilenet} show the evolved local explanations by E-LIME to explain the prediction of ResNet, DenseNet and MobileNet, respectively.

As can be seen from Fig. \ref{fig:elime_shark_resnet}, Fig. \ref{fig:elime_shark_densenet} and Fig. \ref{fig:elime_shark_mobilenet}, very important interpretable features, i.e., the shapes (and sizes) of the pectoral, dorsal fins and a tail, are successfully extracted in the local explanations for interpreting the predictions of ResNet, DenseNet and MobileNet for the image. As we know, the crescent-shaped tail is considered an important feature that could distinguish sharks from many other animals, and the tail size is usually much larger than that of other oceanic fishes. Moreover, E-LIME is able to effectively extract interpretable features, such as, the mouth and jaws, and the contour of the head. All of these explanations could intuitively reveal why this image is predicted into the shark category by ResNet, DenseNet and MobileNet. Moreover, the bubbles are also discerned from the contour of the shark by the learned explanations in Fig. \ref{fig:elime_shark_resnet}, Fig. \ref{fig:elime_shark_densenet} and Fig. \ref{fig:elime_shark_mobilenet}. 

\subsection{Comparison with LIME}

In LIME \cite{ribeiro2016should}, one parameter that needs to be set is the number of interpretable features selected. However, in the proposed method, the number of interpretable features is automatically evolved. To perform a fair comparison, the number of interpretable features in LIME is set as the number evolved by the proposed method. Additionally, both E-LIME and LIME use the same superpixel partitioning generated by the SLIC algorithm for each input image, ensuring a consistent basis for comparison. Since Section \ref{SSS:elime_result_evolved_explanations} has demonstrated that the proposed method can produce effective local explanations for all of the three deep CNN models, the local explanations of ResNet produced by LIME and the proposed method are compared to avoid the content redundancy in this paper. 

\begin{figure}[ht]
	\centering
	\begin{subfloat}[E-LIME]{
			\includegraphics[width=0.22\textwidth]{elime_dog_exp}
			\label{fig:elime_vs_lime_dog_elime}
		}
	\end{subfloat}
	\begin{subfloat}[LIME]{
			\includegraphics[width=0.22\textwidth, height=0.16\textwidth]{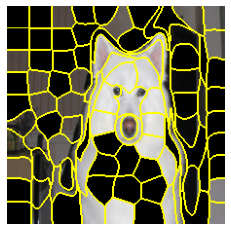}
			\label{fig:elime_vs_lime_dog_lime}
		}
	\end{subfloat}
	\caption{The local explanation comparison between E-LIME and LIME on a Samoyed image.}
	\label{fig:elime_vs_lime_dog}
\end{figure}

It can be observed from Fig. \ref{fig:elime_vs_lime_dog_lime} and Fig. \ref{fig:elime_vs_lime_dog_elime}, both LIME and E-LIME are able to select the important features of the Samoyed because the distinct ears, the main body and the eyes are all displayed in both of the local explanations. Furthermore, the vase of the flower and the hanger in the background that could possibly confuse the deep CNN model to make the prediction, do not appear, so both LIME and E-LIME can filter out the noisy background. 

\begin{figure}[ht]
	\centering
	\begin{subfloat}[E-LIME]{
			\includegraphics[width=0.22\textwidth]{elime_cat_exp}
			\label{fig:elime_vs_lime_cat_elime}
		}
	\end{subfloat}
	\begin{subfloat}[LIME]{
			\includegraphics[width=0.22\textwidth, height=0.17\textwidth]{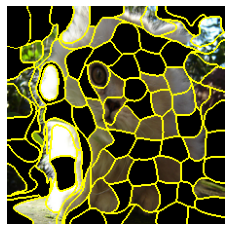}
			\label{fig:elime_vs_lime_cat_lime}
		}
	\end{subfloat}
	\caption{The local explanation comparison between E-LIME and LIME on a Persian cat.}
	\label{fig:elime_vs_lime_cat}
\end{figure}

Similarly to the local explanations of the Samoyed image, the critical interpretable features are identified by both LIME and E-LIME. In Fig. \ref{fig:elime_vs_lime_cat_elime}, the right eye and the two ears, which are decisive to classify the image to a Persian cat, are selected by E-LIME. In Fig. \ref{fig:elime_vs_lime_cat_lime}, two essential features---the left eye and the nose, show up in the local explanation obtained by LIME. E-LIME seems better than LIME because for the same number of selected superpixels, E-LIME includes more superpixels of the cat, while LIME selects more background. As expected, the background noises, such as the leaves next to the left jaw of the Persian cat, are discarded by both LIME and E-LIME. 

\begin{figure}[ht]
	\centering
	\begin{subfloat}[E-LIME]{
			\includegraphics[width=0.22\textwidth]{elime_eagle_exp}
			\label{fig:elime_vs_lime_eagle_elime}
		}
	\end{subfloat}
	\begin{subfloat}[LIME]{
			\includegraphics[width=0.22\textwidth, height=0.18\textwidth]{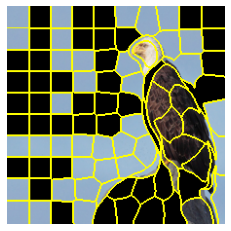}
			\label{fig:elime_vs_lime_eagle_lime}
		}
	\end{subfloat}
	\caption{The local explanation comparison between E-LIME and LIME on a bald eagle.}
	\label{fig:elime_vs_lime_eagle}
\end{figure}

\begin{figure}[ht]
	\centering
	\begin{subfloat}[E-LIME]{
			\includegraphics[width=0.22\textwidth]{elime_shark_exp}
			\label{fig:elime_vs_lime_shark_elime}
		}
	\end{subfloat}
	\begin{subfloat}[LIME]{
			\includegraphics[width=0.22\textwidth, height=0.15\textwidth]{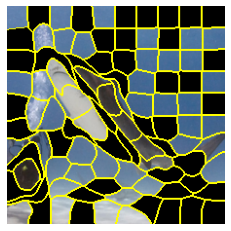}
			\label{fig:elime_vs_lime_shark_lime}
		}
	\end{subfloat}
	\caption{The local explanation comparison between E-LIME and LIME on a tiger shark.}
	\label{fig:elime_vs_lime_shark}
\end{figure}

The similar pattern of comparing the local explanations produced by LIME and E-LIME can be noticed in Fig. \ref{fig:elime_vs_lime_eagle} and Fig. \ref{fig:elime_vs_lime_shark}. For the bald eagle image, important features, such as the head, the hooked beak, and the brown wings of the bald eagle, are present in the local explanations of LIME and E-LIME. Most parts of the rock, especially the parts that are next to the foot and wings of the bald eagle, are taken out by both LIME and E-LIME. Regarding the tiger shark image, the unique features for a tiger shark are picked by both LIME and E-LIME, e.g. the pectoral, dorsal fins, and the tail, even though different parts of the tail show up in the local explanations of LIME and E-LIME. It can be noted that the diver and the other fish, which could distract the deep CNN model to predict a tiger shark, are wiped out by both LIME and E-LIME. 

To sum up, E-LIME can evolve local explanations that are very competitive to those obtained by LIME. However, as mentioned earlier, the proposed targets are to reduce the computational cost and eliminate the human intervention of fine-tuning the number of interpretable features by keeping the quality and explainability of the local explanations. 

\subsection{Further Experiment on Vision Transformer}

An additional experiment has been conducted to explain the predictions made by the recently introduced Vision Transformer \cite{dosovitskiy2020image}. Vision Transformer is a powerful method recently proposed for computer vision tasks, which was originally designed for natural language processing. The proposed method was evaluated to explain the Vision Transformer, which differs from other Deep Convolutional Neural Networks. The results with Vision Transformer are presented in this sub-section.

\begin{figure}[ht] 
	\centering
	\begin{subfloat}[Original image]{
			\includegraphics[width=0.22\textwidth]{elime_dog_orig}
			\label{fig:molime_dog_orig_vit}
		}
	\end{subfloat}
	\begin{subfloat}[1st evolved local explanation for Vision Transformer]{
			\includegraphics[width=0.22\textwidth]{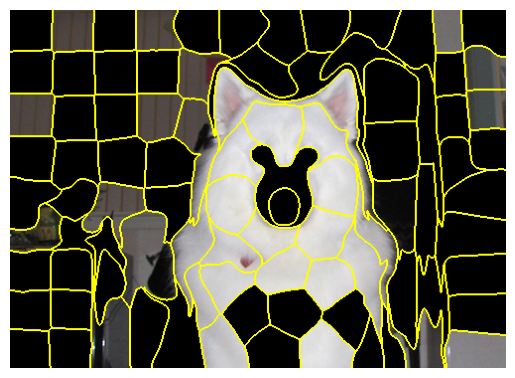}
			\label{fig:elime_dog_exp_vit}
		}
	\end{subfloat}

	\caption{Evolved local explanations for the image of a Samoyed based on \textbf{Vision Transformer}.}
	\label{fig:elime_dog_vit}
\end{figure}

A local explanation shown in Figure \ref{fig:elime_dog_vit} is evolved by E-LIME to help humans understand why Vision Transformer makes a particular prediction about an image of a Samoyed. The explanation highlights the shape and size of the Samoyed's ears as the most informative features, which distinguish Samoyeds from other breeds. Interestingly, the features that contain the eyes and the nose of the Samoyed have been masked out in the local explanation, which means Vision Transformer does not rely on the eyes and the nose to make the prediction. It may be because some other breeds have similar features. The local explanation also filters out potentially noisy features such as background objects, including a vase of flowers, a cabinet and a hanger.

\begin{figure}[ht] 
	\centering
	\begin{subfloat}[Original image]{
			\includegraphics[width=0.22\textwidth]{elime_cat_orig}
			\label{fig:elime_cat_orig_vit}
		}
	\end{subfloat}
	\begin{subfloat}[1st evolved local explanation for Vision Transformer]{
			\includegraphics[width=0.22\textwidth]{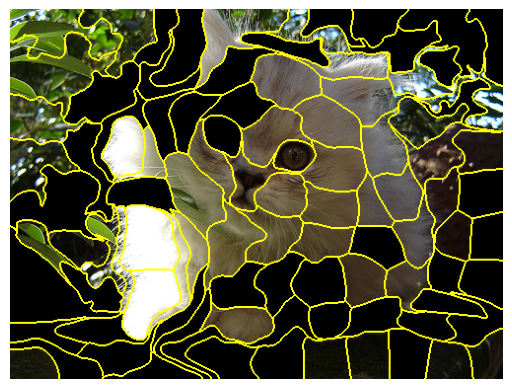}
			\label{fig:elime_cat_exp_vit_1}
		}
	\end{subfloat}

	\caption{Evolved local explanations for the image of a Persian cat based on \textbf{Vision Transformer}.}
	\label{fig:elime_cat_vit}
\end{figure}

Figure \ref{fig:elime_cat_vit} demonstrates that the proposed method has successfully learned a local explanation for Vision Transformer. It presents informative and meaningful features, such as the ears, nose, mouth, and right eye of the Persian cat, which are essential for identifying the object as a cat. Notably, two green leaves in the image are in close proximity to the left lower jaw of the Persian cat, but the local explanation has not selected them as they are not relevant to the cat's recognition. The interpretable features explain the classification outcome with regard to the Persian cat.

To sum up, the proposed method has the ability to learn local explanations that explain why the Vision Transformer makes predictions on the two selected images - a Samoyed dog and a Persian cat.

\subsection{Performance Analysis}
\begin{table}
	\centering
	\caption{Results - \textbf{ResNet}.}
	\label{tb:result_exp_fitness_resnet}
	\renewcommand\arraystretch{1.6}
	\renewcommand\tabcolsep{2pt} 
	\scriptsize
	\begin{tabular}{|c|c|cc|cc|} 
		\hline
		\multicolumn{1}{|c|}{} & \textbf{\tabincell{c}{Probability}} &\multicolumn{2}{c|}{\textbf{\tabincell{c}{Fitness \\ (probability)}} } &\multicolumn{2}{c|}{\textbf{\tabincell{c}{Training time \\(Seconds)}}}   \\ \hline 
		
		\textbf{Images}	 & \textbf{Original} &\textbf{Best} &\tabincell{c}{\textbf{Mean} $\pm$ \textbf{Std}} &\textbf{Shortest}&\tabincell{c}{\textbf{Mean} $\pm$ \textbf{Std}}\\ \hline 
		
		Samoyed & 0.9574 &0.9995  &0.9988 $\pm$ 0.0004  & 36.3771& 38.0843 $\pm$ 0.9009 \\ \hline
		
		Persian cat & 0.9793& 0.9994  &0.9981 $\pm$0.0013  & 33.3398 &  35.3640$\pm$  0.9839 \\ \hline
		
		bald eagle  &  0.9851 &0.99996  &  0.99991$\pm$  0.00004 &38.0997 &  41.8636 $\pm$  1.2395\\ \hline
		tiger shark  & 0.8876& 0.9903 &0.9775 $\pm$0.0113  & 43.0951 & 45.7559  $\pm$1.7464 \\ \hline				
	\end{tabular}%
	\begin{tablenotes}
		\item 1: Std means standard deviation.
	\end{tablenotes}
\end{table}

\begin{table}
	\centering
	\caption{Results - \textbf{DenseNet}.}
	\label{tb:result_exp_fitness_densenet}
	
	\renewcommand\arraystretch{1.6}
	\renewcommand\tabcolsep{1pt} 
	\scriptsize
	\begin{tabular}{|c|c|cc|cc|} 
		\hline
		\multicolumn{1}{|c|}{} & \textbf{\tabincell{c}{Probability}} &\multicolumn{2}{c|}{\textbf{\tabincell{c}{Fitness \\ (probability)}} } &\multicolumn{2}{c|}{\textbf{\tabincell{c}{Training time \\(Seconds)}}}   \\ \hline 
		
		\textbf{Images}	 & \textbf{Original} &\textbf{Best} &\tabincell{c}{\textbf{Mean} $\pm$ \textbf{Std}} &\textbf{Shortest}&\tabincell{c}{\textbf{Mean} $\pm$ \textbf{Std}}\\ \hline

		Samoyed & 0.9196 &0.9898  &0.9832 $\pm$ 0.0047  & 39.6621& 41.9406 $\pm$ 1.1387 \\ \hline
		
		Persian cat & 0.9937 & 0.9998  &0.9994 $\pm$0.0003  & 39.2329 &  41.3979$\pm$  1.1005 \\ \hline
		
		bald eagle &  0.9956 &0.999988  &  0.999979$\pm$  0.000007 &45.3849&  47.5821 $\pm$  0.9506\\ \hline
		tiger shark & 0.7970 & 0.9825 &0.9735 $\pm$0.0065  & 46.9843 & 51.122737 $\pm$1.8917 \\ \hline				
	\end{tabular}%
	\begin{tablenotes}
		\item 1: Std means standard deviation.
	\end{tablenotes}
\end{table}

\begin{table}
	\centering
	\caption{Results - \textbf{MobileNet}.}
	\label{tb:result_exp_fitness_mobilenet}
	\renewcommand\arraystretch{1.6}
	\renewcommand\tabcolsep{1pt} 
	\scriptsize
	\begin{tabular}{|c|c|cc|cc|} 
		\hline
		\multicolumn{1}{|c|}{} & \textbf{\tabincell{c}{Probability}} &\multicolumn{2}{c|}{\textbf{\tabincell{c}{Fitness \\ (probability)}} } &\multicolumn{2}{c|}{\textbf{\tabincell{c}{Training time \\(Seconds)}}}   \\ \hline 
		
		\textbf{Images}	 & \textbf{Original} &\textbf{Best} &\tabincell{c}{\textbf{Mean} $\pm$ \textbf{Std}} &\textbf{Shortest}&\tabincell{c}{\textbf{Mean} $\pm$ \textbf{Std}}\\ \hline 
		
		Samoyed  & 0.5186 & 0.9975 &0.9945 $\pm$0.0019  & 34.5222 & 36.2646  $\pm$0.8991 \\ \hline	
		
		Persian cat  & 0.9914 & 0.9996  &0.9998 $\pm$0.0001  & 33.1101 &  35.4532$\pm$  1.0641 \\ \hline
		
		bald eagle &  0.9573 &0.999993  &  0.999982$\pm$  0.000007 &37.8212 &  39.7936 $\pm$  1.0737\\ \hline
		tiger shark  & 0.9377  & 0.9903 &0.9800 $\pm$0.0067  & 40.1384 & 45.4999  $\pm$2.1441 \\ \hline				
	\end{tabular}%
	\begin{tablenotes}
		\item 1: Std means standard deviation.
	\end{tablenotes}
\end{table} 

In TABLE \ref{tb:result_exp_fitness_resnet}, the second column shows the original probabilities achieved by ResNet on the four original images, and the third column reports the mean probabilities of the 30 runs achieved by the best individual from the final generation in E-LIME. By comparing between the original probabilities and mean probabilities in TABLE \ref{tb:result_exp_fitness_resnet}, E-LIME improves the probabilities on all the four images, and thereby increases the confidence of users when determining whether these predictions are trustworthy or not. More specifically, on the two images (i.e., Samoyed and tiger shark), compared with original probabilities, the mean probabilities (achieved by E-LIME) increase by 0.1344 (i.e., 13.44\%) and 0.0286 (i.e., 2.86\%), respectively. The final column in TABLE \ref{tb:result_exp_fitness_resnet} reports the training time of E-LIME, which shows the proposed E-LIME method is much more time-efficient than LIME which uses \textbf{10 minutes}.

TABLE \ref{tb:result_exp_fitness_densenet} and TABLE \ref{tb:result_exp_fitness_mobilenet} report the results of E-LIME for interpreting predictions of DenseNet and MobileNet, respectively, on the four images. As can be seen from TABLE \ref{tb:result_exp_fitness_densenet}, the mean probabilities (of the 30 runs) achieved by E-LIME are better than the original probabilities of DenseNet on all the four images, particularly on the images of a Samoyed (increased by 3.36\%) and a tiger shark (increased by 17.76\%). 


Similarly, according to the mean probabilities shown in TABLE \ref{tb:result_exp_fitness_mobilenet}, E-LIME can also improve the original probabilities of MobileNet on the four images, particularly on the images of a Samoyed (improved by 47.59\%), a bald eagle (improved by 4.10\%), and a tiger shark (improved by 4.23\%). The improved probability on each image can increase the confidence of users when they need to judge whether these predictions are trustworthy or not. Based on the last columns in TABLE \ref{tb:result_exp_fitness_densenet} and TABLE \ref{tb:result_exp_fitness_mobilenet}, the training time of E-LIME for interpreting the predictions of DenseNet and MobileNet is much faster than LIME.  

Besides, as can be seen from TABLE \ref{tb:result_exp_fitness_resnet}, TABLE \ref{tb:result_exp_fitness_densenet} and TABLE \ref{tb:result_exp_fitness_mobilenet}, by using E-LIME on the four images, the standard deviation of the 30 probabilities are very small, and the gap between the best and mean probabilities is also very narrow. This is mainly because E-LIME is able to effectively extract the informative features in the local explanations to improve the confidence of a prediction. 

By comparing the results in TABLE \ref{tb:result_exp_fitness_resnet}, TABLE \ref{tb:result_exp_fitness_densenet} and TABLE \ref{tb:result_exp_fitness_mobilenet}, it is noticed that mean probabilities achieved by E-LIME on the four images for the three different deep CNNs are very similar. Furthermore, the training time of E-LIME based on the three deep CNNs are also similar. This could further show that the proposed method is able to effectively and efficiently evolve the model-agnostic explanations. 

\section{Conclusions and Future Work} \label{SSS:elime_conclusion}

The overall goal of proposing a novel GA-based method to evolve local interpretable model-agnostic explanations has been achieved. The proposed method has shown its effectiveness and efficiency from the experimental results. Firstly, inspired by LIME, the interpretable features, i.e., the superpixels, have been adopted in the proposed method, which are encoded in a binary vector. GA can be easily applied to evolve that and optimise it. Secondly, by applying GA to achieve the local explanations instead of a linear model with Lasso regularisation in LIME, the time-consuming sampling process has been eliminated, so the proposed method requires much less computational cost than LIME. Lastly, a new fitness function has been proposed, which takes the probability of the deep CNN model to predict a specific label of the image as the fitness value. As the probability represents the confidence of the deep CNN model to make a specific prediction, the target of proposed method is to search for an optimal subset of interpretable features that boosts the confidence of deep CNN model. Therefore, the users can examine the evolved subset to figure out the reasons behind the predictions.

A limitation of this paper is that no quantitative metrics were proposed. In future work, we plan to explore the development of quantitative metrics, such as faithfulness or stability, to evaluate the interpretability of local explanations in a more rigorous way. Additionally, it would be beneficial to investigate methods for selecting a subset of images that effectively represent the entire dataset, enabling the generation of global explanations. Furthermore, the proposed method currently relies on SLIC for generating superpixels, which may not perform well on texture or binary images. Exploring more sophisticated superpixel generation techniques could lead to improved outcomes. Last but not least, the research focuses on deep convolutional neural networks for image classification tasks. It is important to extend the scope of the machine learning task from supervised classification to semi-supervised learning or even unsupervised learning, and also extend the scope of data type from image to text or audio.

\ifCLASSOPTIONcaptionsoff
  \newpage
\fi



\bibliographystyle{IEEEtran}
\bibliography{elime_tevc}
\newpage




\end{document}